\documentclass{article}
\usepackage{cite}
\usepackage{spconf,amsmath,amssymb,amsfonts}
\usepackage{algorithmic}
\usepackage{graphicx}
\usepackage{textcomp}
\usepackage{multirow}
\usepackage{adjustbox}

\usepackage[caption=false,font=scriptsize]{subfig}

\usepackage{stfloats}
\usepackage[table,xcdraw]{xcolor}
\usepackage[colorlinks,citecolor=red,urlcolor=blue,bookmarks=false,hypertexnames=true,pagebackref=true]{hyperref} 

\def\BibTeX{{\rm B\kern-.05em{\sc i\kern-.025em b}\kern-.08em
    T\kern-.1667em\lower.7ex\hbox{E}\kern-.125emX}}

\title{SYMMETRIC CONVOLUTIONAL FILTERS: A NOVEL WAY TO CONSTRAIN PARAMETERS IN CNN}
\name{Harish Agrawal, Sumana T., S.K. Nandy}
\address{CAD Lab, Indian Institute of Science, Bengaluru, India 560012\\
Email:(harisha,sumanat,nandy)@iisc.ac.in}
\begin{document}
%\ninept
%
\maketitle
\begin{abstract}
We propose a novel technique to constrain parameters in CNN based on symmetric filters. We investigate the impact on SOTA networks when varying the combinations of symmetricity. We demonstrate that our models offer effective generalisation and a structured elimination of redundancy in parameters. We conclude by comparing our method with other pruning techniques.
\end{abstract}
\section{Introduction}
\label{sec:intro}
Neural Networks (NNs) have been successful compared to other Machine Learning (ML) techniques, mainly due to their ability to extract representative features from the intrinsic structure of raw input data. Moreover, the emergence of smart Internet of Things (IoT) endpoints and social media have opened the floodgates for colossal amounts of data. Combined with the increasing availability of massive data sets, NNs are espoused to be part and parcel of futuristic Artificial Intelligence (AI) systems and products. However, the SOTA NN models are over-parameterised \cite{PredictingParams} and have huge computational requirements that have doubled every few months \cite{GreenAI-blog}. For deployment in edge devices such as smartphones and wearables, the heavily parameterised SOTA networks should be constrained and compressed to lower their memory footprint and computational cost \cite{PredictingParams}.

NN model compression methods include not only techniques for pruning \cite{PruningEfficientCNN,HRank,NISP} or quantizing parameters \cite{BinarizedNN} of a trained model, but also designing models that are optimal in parameters \cite{DepthwiseSepConv, MobileNetV2}. We observe that the proliferation of deep CNN models has matured along two directions as, \textit{standard} networks along improving accuracy of the model in general and \textit{edge} networks along explicitly targetting inference in edge devices while maintaining acceptable accuracy. The \textit{standard} networks \cite{ResNet, GoogleNet, DenseNet} employ standard convolutions in their basic modules to create deeper networks. While the \textit{edge} networks \cite{MobileNet, MobileNetV2} replace the expensive, standard convolutions with depthwise (DW) convolutions. Also, they sandwich DW convolutions between pointwise convolution ($1\times1$), forming bottleneck blocks to control the number of channels fed into them, thereby maintaining the parameters in the network.

\begin{figure}[h!]
\centering
\includegraphics[width=1.5in]{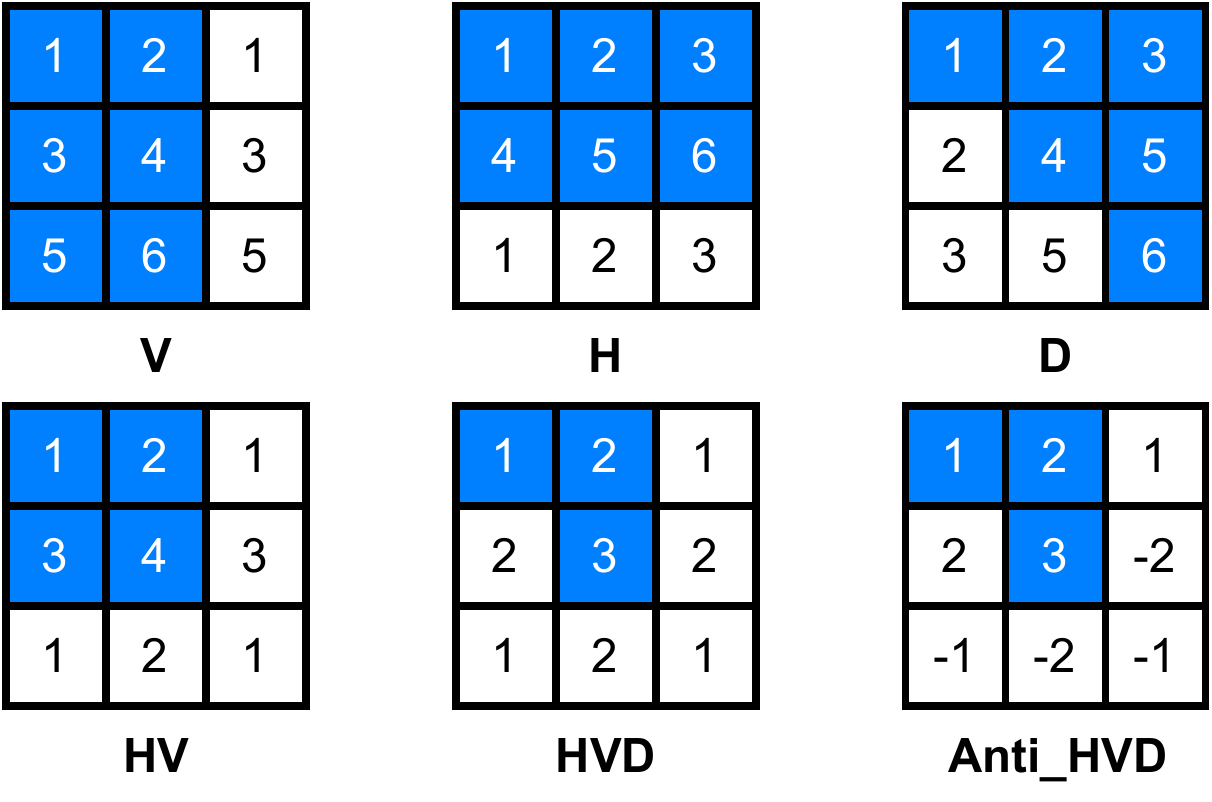}
\caption{\small Various types of symmetric filters; V: Vertically Symmetric, H: Horizontally Symmetric, D: Diagonally Symmetric, HV: Horizontally and Vertically Symmetric, HVD: Horizontally, Vertically and Diagonally Symmetric , Anti\_HVD: Same as HVD but with negative coefficients; Compared to a standard $3\times3$ filter: V,H,D use 33\% less parameters and HVD,Anti\_HVD use 66\% less parameters;}
\label{fig:DiffSymmFilters}
\end{figure}

In this work, we propose a novel NN model compression technique using Symmetric Convolutional Filters (SCF). Standard filters such as Gaussian smoothing, Laplacian edge detection, box blur, sharpen are used in traditional image processing pipelines irrespective of the target application. Their 2D kernels mostly exhibit symmetricity about at least one axis. For example, filter weights mirrored about their central vertical axis are vertically symmetric; about their central horizontal axis are horizontally symmetric; about their diagonal are diagonally symmetric. Figure \ref{fig:DiffSymmFilters} illustrates a few symmetric kernels in which blue background indicates free coefficients and the rest are tied to their symmetric counterparts. We use these 2D kernels for SCF in our work, where free coefficients are the only trainable parameters. 

Using SCF instead of standard convolutional filters in several \textit{standard} networks, it's possible to reduce their parameters comparable to an \textit{edge} network without significantly affecting their accuracy. Moreover, the structural properties in SCF can be leveraged by hardware accelerators to reduce MAC computations (by pre-adding the inputs before multiplying them with corresponding weights), leading to an efficient edge implementation. Although, in this work we do not discuss the implementation issues inherent to NN architectures (memory consumption, FLOPS, amenability to acceleration, etc). Prior works \cite{symnet, stable-symnet} that have explored symmetric constraints on CNN weights are limited to MNIST and/or smaller custom networks. To the best of our knowledge, we believe our work is the first to empirically validate SCF in SOTA deep networks with real world image datasets.

The other commonly used technique in NN model compression, network pruning, involves removing redundant parameters that have a negligible effect on output accuracy depending on various metrics. Generally, pruning is an iterative process involving repetitive pruning and tuning pre-trained models under heavy supervision. In contrast, SCF networks are trained in the usual way, similar to the \textit{standard} networks. Moreover, pruning and quantizing NN weights are complementary to our approach. If required, a trained NN model with symmetric filters can later be quantized and/or pruned for further model compression.

In this paper, we empirically validate model compression of SOTA networks using extensive combinations of symmetricity. We show that our SCF models show less overfitting compared to heavily over-parameterised, less compact base models. We extend the exploration to compact edge networks. Finally, we compare our results with other pruning techniques. Our concluding remarks from the overall experimental results are the following. Certain dimensions are absolutely redundant for representation in the visual domain. Therefore, the intention should not be about just reducing the number of parameters of an over-parameterised CNN model but should be about finding those absolute redundant dimensions. The evolution of NN architecture from perceptron models, which were fully connected networks, to convolutional models, led to eliminating some of those absolute redundant dimensions. Similarly, we infer that models with SCF eliminate a few more of those redundant dimensions. With this insight, we believe employing SCF in future exploration of NN models is imminent.

%The rest of the paper is organised as follows. In Section\ref{sec:motivation}, we discuss the motivation for symmetric convolutional filters from neuron science. Next, in Section \ref{sec:relatedWork} we compare our work with other CNN compression techniques. Finally in Section \ref{sec:exp}, we discuss the extensive experiments and results.% that we have carried out to explore the different combinations and levels of symmetricity.

\section{Motivation from Neuron Science}
\label{sec:motivation}
The roots of AI and its inherent progress has very much been influenced from the structure and the method of learning in human brain. The basis of the hierarchical NN structure that we find today in all deep NN is from the hierarchy structure of a visual cortex \cite{NeoCognitron}. As we go deeper into the hierarchy in a NN, the field of view of a neuron increases, allowing it to see bigger parts of the image. Consequently, NN models transform raw input data into higher dimensional feature space through the hierarchy of layers. Visualisation of features learnt by the network reveals that the initial layers learn simple features such as edges, colors and textures, whereas the later layers learn more complex and task-specific features \cite{deconv_viz}.

The fMRI based study conducted in \cite{fMRI_BrainStudy} focuses on the neural activity in the following visual areas of the human brain: \textit{early (V1, V2, V3), dorsal, lateral, temporal, ventral} and indicates that visual areas starting from \textit{V3} and all the way down until \textit{ventral} fire on visual stimuli containing symmetric patterns. Their observations conclude that the neural responses for tasks involving symmetry detection were not only more prominent when compared to passive viewing, but also were proportional to the percentage of symmetricity.

\begin{table}[]
\centering
\resizebox{\columnwidth}{1.5cm}{
\tiny
\begin{tabular}{ll|l|c|c}
\hline
\multicolumn{2}{c|}{\textbf{Name}}                                       & \multicolumn{1}{c|}{\textbf{\begin{tabular}[c]{@{}c@{}}Filters\\ Used\end{tabular}}} & \textbf{\begin{tabular}[c]{@{}c@{}}Test \\ Error \%\end{tabular}} & \textbf{\begin{tabular}[c]{@{}c@{}}\% Original\\ Params\end{tabular}} \\ \hline
\rowcolor[HTML]{EFEFEF} 
\multicolumn{2}{l|}{\cellcolor[HTML]{EFEFEF}\textbf{Base Model}}         & 64 - Standard                                                                        & 4.4                                                               & 100                                                                   \\ \hline
                  & \textbf{C}                                           & 64 - V                                                                               & 4.45                                                              & 66.66                                                                 \\
\rowcolor[HTML]{EFEFEF} 
\textbf{Type-III} & \textbf{B}                                           & 64 - H                                                                               & 4.89                                                              & 66.66                                                                 \\
\textbf{}         & \textbf{A}                                           & 32-H, 32-V                                                                           & 4.32                                                              & 66.66                                                                 \\ \hline
\rowcolor[HTML]{EFEFEF} 
\textbf{}         & \textbf{C}                                           & 64 - HVD                                                                             & 4.52                                                              & 33.33                                                                 \\
\textbf{Type-II}  & \textbf{B}                                           & 64 - Anti HVD                                                                        & 4.51                                                              & 33.33                                                                 \\
\rowcolor[HTML]{EFEFEF} 
\textbf{}         & \cellcolor[HTML]{EFEFEF}                             & 32 - HVD,                                                                            & \cellcolor[HTML]{EFEFEF}                                          & \cellcolor[HTML]{EFEFEF}                                              \\
\rowcolor[HTML]{EFEFEF} 
\textbf{}         & \multirow{-2}{*}{\cellcolor[HTML]{EFEFEF}\textbf{A}} & 32 - Anti HVD                                                                        & \multirow{-2}{*}{\cellcolor[HTML]{EFEFEF}4.32}                    & \multirow{-2}{*}{\cellcolor[HTML]{EFEFEF}33.33}                       \\ \hline
\textbf{}         &                                                      & 16 - H, 16 - V,                                                                      &                                                                   &                                                                       \\
\textbf{Type-I}   &                                                      & 16 - HVD,                                                                            &                                                                   &                                                                       \\
                  &                                                      & 16 - Anit HVD                                                                        & \multirow{-3}{*}{4.31}                                            & \multirow{-3}{*}{50}                                                  \\ \hline
\end{tabular}
}
\caption{\small Test Error on CIFAR-10 Dataset when only the Standard Convolutional Filters of First Layer of ResNeXt29\_32x4d Model are replaced with different SCF configurations; Refer to Figure \ref{fig:DiffSymmFilters} for Types of Symmetric Filters.}
\label{table:FirstLayerExperiment}
\end{table}

In order to imbibe the \textit{higher neuronal excitation for symmetric features} trait observed in brain \cite{fMRI_BrainStudy}, in our CNN model, we propose structurally constraining the convolutional filters with symmetric filters. In CNN, the neurons that show high response for certain features need to have weights similar to the features in the transformed space. To enhance the ability to detect symmetry, we can enforce the neurons to retain the spatial structure of the features even in higher dimensions. Therefore, applying symmetric constraints on standard convolution filters makes sense.

\section{Comparison with related work in NN Model Compression}
\label{sec:relatedWork} 
MobileNet \cite{MobileNet} uses depthwise separable convolutions \cite{DepthwiseSepConv}, which is a form of factorised convolution akin to factorising standard convolution into depthwise convolution and pointwise convolution ($1\times1$ convolution). Similarly, MobileNetV2 \cite{MobileNetV2} also uses depthwise separable convolution but incorporates their novel inverted residual connections with linear bottleneck. When compared with our proposal of symmetric convolution, which is spatially constraining the standard convolution, depthwise separable convolution constrains standard convolution in the depth dimension. Our Experiment 3 (see Section \ref{sec:exp3}) explores constraining filters in both spatial and depth dimensions.

Pruning techniques are classified as structured and unstructured \cite{PruningEfficientCNN}. We compare with other structured pruning techniques because SCF model compression is more akin to structured method as they do not introduce random sparsity in the filter kernels. Li et al. \cite{PruningEfficientCNN} proposed pruning those filters from CNN  by using the sum of absolute weight of the filters. The idea behind HRank \cite{HRank} is that low-rank feature maps contain less information and thus can be used to find unimportant filters in CNN. NISP \cite{NISP} applies the feature ranking technique to measure the importance of each neuron and then formulate network pruning as a binary integer optimisation problem. See Section \ref{sec:exp_prune} for comparison with pruning methods.

\begin{figure*}[t!]
\centering
\subfloat[CIFAR-10 Validation Error vs \#Parameters] 
{\includegraphics[width=2.3in, height=2in]{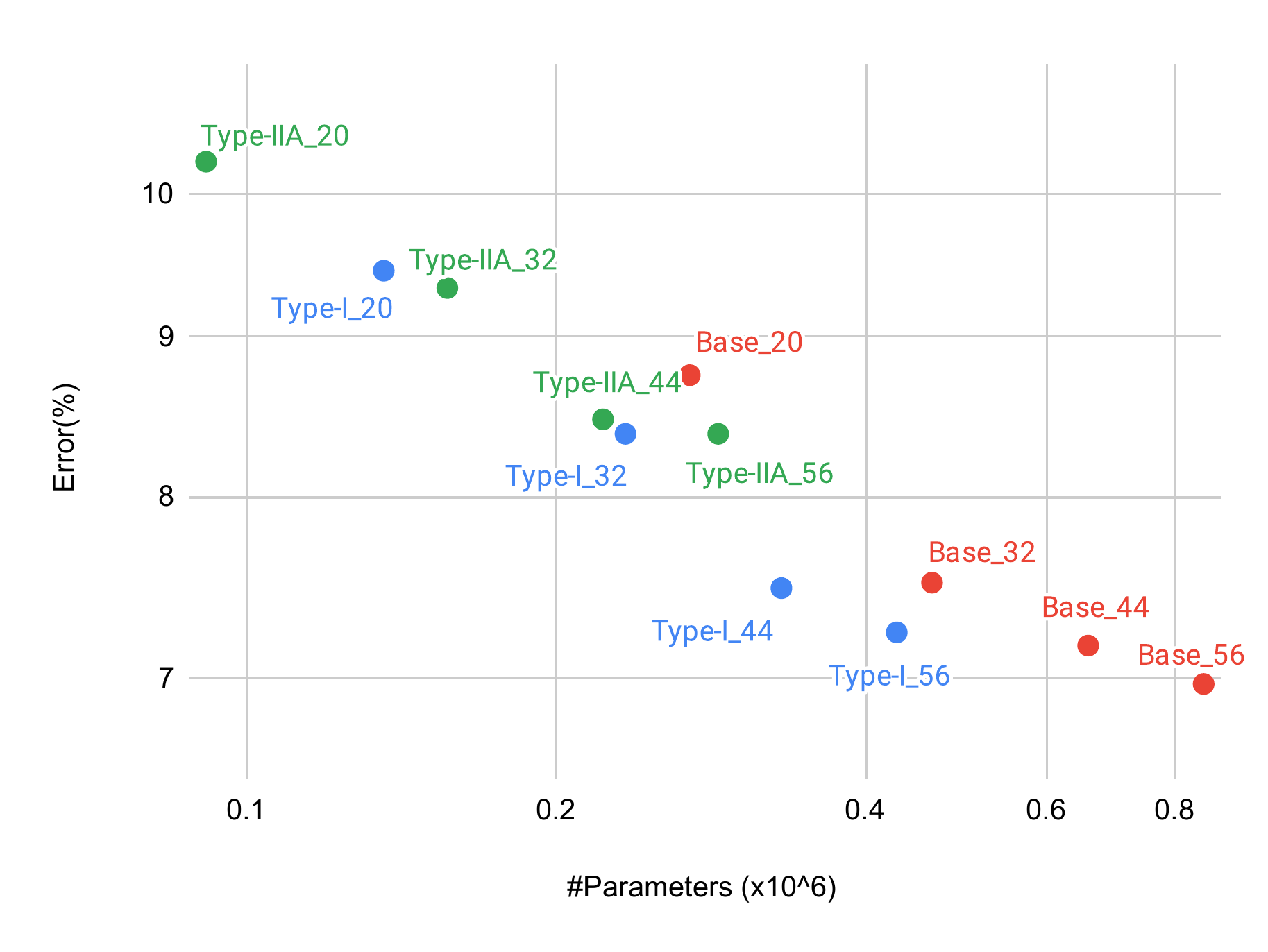}
\label{fig:validationErrorVsParamResNet:first}}
\hfil
\subfloat[CIFAR-100 Validation Error vs \#Parameters] {\includegraphics[width=2.3in, height=2in]{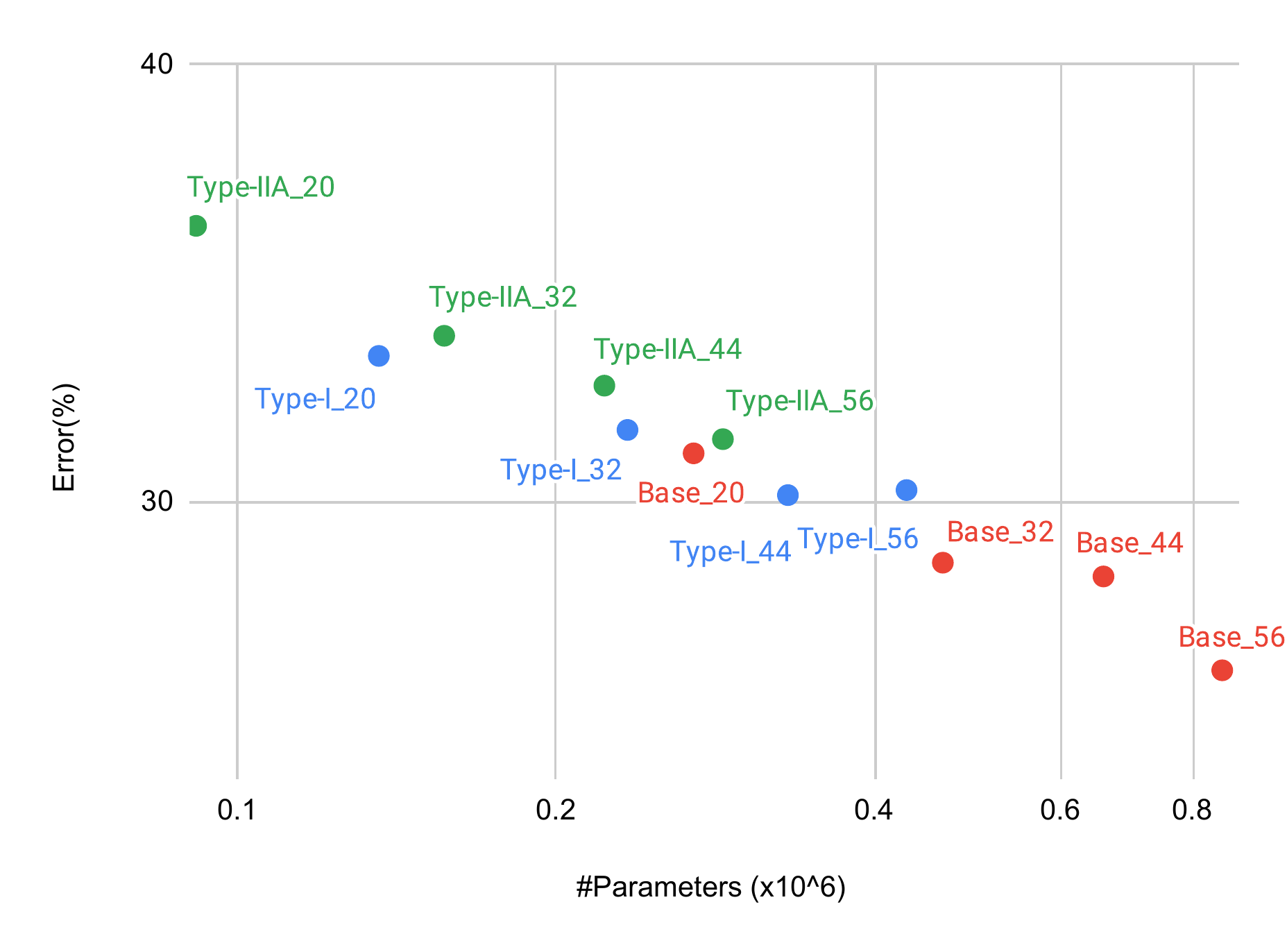}
\label{fig:validationErrorVsParamResNet:sec}}
\hfil
\subfloat[Training and Validation Error vs \#epochs on CIFAR-100]{\includegraphics[width=2.3in,
height=2in]{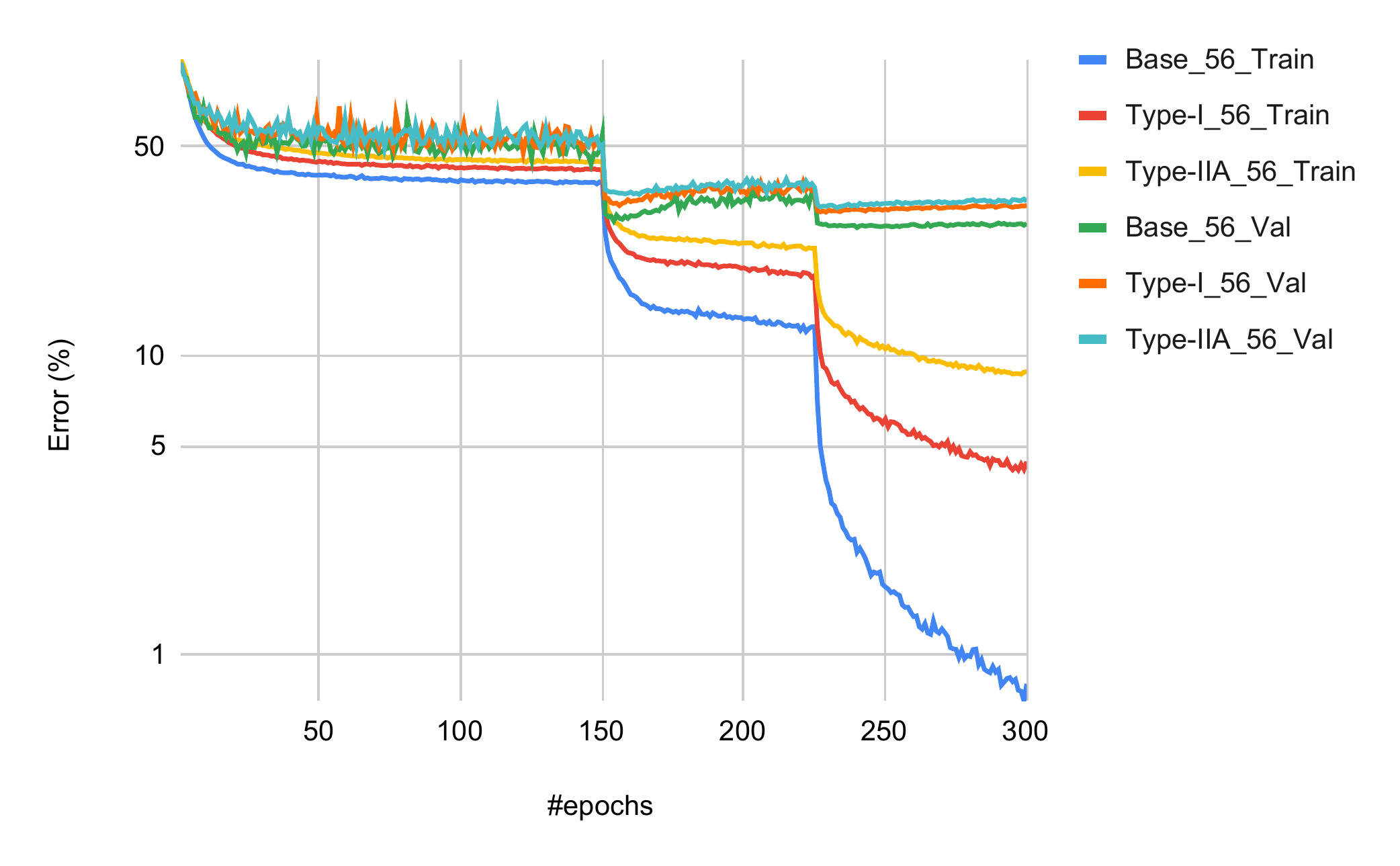}
%\caption{\small Training and Validation Error(\%) of base ResNet-56, Type-I and Type-IIA ResNet-56 on CIFAR-100 dataset.}
\label{fig:Train_Val_AccuracyResNet56}}
\caption{\small Training and Validation Error of Base and SCF ResNet models with 20, 32, 44 and 56 depth on different datasets}
\label{fig:validationErrorVsParamResNet}
\end{figure*}

\section{Experimental Setup and Results}
\label{sec:exp}
\subsection{Experimental Setup}
For all our experiments, we have used the NVIDIA DGX-1 system. It has a total of 8 NVIDIA Tesla V100 GPUs, out of which we have used only 1 with 32 GB GPU memory. All code is written in PyTorch\footnote{Code will be released soon on Github}. We have experimented with CIFAR-10 and CIFAR-100 datasets \cite{CIFAR}. CIFAR-10 dataset consists of 60,000 32x32 real-world colour images of 10 classes, each class having 6,000 images. CIFAR-100 dataset is very similar to the CIFAR-10 dataset, except it has 100 categories containing 600 images each. Both datasets are divided into 50,000 images as the training set and the remaining 10,000 images as the test set. We have used momentum based mini-batch gradient descent \cite{SGD} algorithm with a batch-size of 128 and momentum of 0.9 with weight-decay of 0.0005. All the configurations are trained for 300 epochs similar to \cite{ResNeXt}, with initial learning rate of 0.1, reduced to 0.01 after 150 epochs, and further reduced to 0.001 after 225 epochs. The input image is $32\times32$ randomly cropped from a zero-padded 40×40 image or its flipping \cite{ResNet}. We have used kaiming normal \cite{KaimingInit} initialisation to initialise weights of all the models trained in our experiments. We have reported best of 3 runs results using Top-1 accuracy. All the networks without SCF are respectively referred as Base models.

In order to find the best combination of SCF configurations to adapt for later experiments, we replace the standard convolutional filters from only the first convolutional layer which directly interacts with raw data, the image itself instead of feature maps, with symmetric configurations as listed in Table \ref{table:FirstLayerExperiment}. We randomly choose ResNeXt29\_32x4d \cite{ResNeXt} model which has $64$ filters of $3\times3\times3$ dimension in its first layer and train it on CIFAR-10 dataset. From the Table \ref{table:FirstLayerExperiment}, Type-I or Type-IIA symmetric configurations provide higher compression for similar accuracy. Hence, we limit the rest of the experiments to SCF configurations with either Type-I or Type-IIA symmetric filters.

\begin{table}[b]
\tiny
\centering
\resizebox{\columnwidth}{1.4cm}{
\begin{tabular}{l|l|cc|cc|c}
\hline
\multicolumn{1}{c|}{}                                                                               & \multicolumn{1}{c|}{}                                                                                  & \multicolumn{2}{c|}{\textbf{\begin{tabular}[c]{@{}c@{}}CIFAR-10\\ Test Error\end{tabular}}}                                                            & \multicolumn{2}{c|}{\textbf{\begin{tabular}[c]{@{}c@{}}CIFAR-100\\ Test Error\end{tabular}}}                                                           &                                                                                                 \\ \cline{3-6}
\multicolumn{1}{c|}{\multirow{-2}{*}{\textbf{\begin{tabular}[c]{@{}c@{}}CNN\\ Model\end{tabular}}}} & \multicolumn{1}{c|}{\multirow{-2}{*}{\textbf{\begin{tabular}[c]{@{}c@{}}Filter\\ Conf.\end{tabular}}}} & \multicolumn{1}{c|}{\textbf{\begin{tabular}[c]{@{}c@{}}Error\\ \%\end{tabular}}} & \textbf{\begin{tabular}[c]{@{}c@{}}\% Inc.\\ wrt Base\end{tabular}} & \multicolumn{1}{c|}{\textbf{\begin{tabular}[c]{@{}c@{}}Error\\ \%\end{tabular}}} & \textbf{\begin{tabular}[c]{@{}c@{}}\% Inc.\\ wrt Base\end{tabular}} & \multirow{-2}{*}{\textbf{\begin{tabular}[c]{@{}c@{}}Parameter or\\ \%Compression\end{tabular}}} \\ \hline
\rowcolor[HTML]{EFEFEF} 
                                                                                                    & \textbf{Base}                                                                                          & \multicolumn{1}{c|}{\cellcolor[HTML]{EFEFEF}4.24}                                & -                                                                   & \multicolumn{1}{c|}{\cellcolor[HTML]{EFEFEF}19.82}                               & -                                                                   & 6.96M                                                                                           \\
\textbf{DenseNet-121}                                                                               & \textbf{Type-I}                                                                                        & \multicolumn{1}{c|}{5.64}                                                        & 1.46\%                                                              & \multicolumn{1}{c|}{21.26}                                                       & \cellcolor[HTML]{FFFFFF}1.80\%                                      & 15.38\%                                                                                         \\
\rowcolor[HTML]{EFEFEF} 
\textbf{}                                                                                           & \textbf{Type-IIA}                                                                                      & \multicolumn{1}{c|}{\cellcolor[HTML]{EFEFEF}5.33}                                & 1.14\%                                                              & \multicolumn{1}{c|}{\cellcolor[HTML]{EFEFEF}22.44}                               & 3.27\%                                                              & 20.51\%                                                                                         \\ \hline
\textbf{}                                                                                           & \textbf{Base}                                                                                          & \multicolumn{1}{c|}{4.59}                                                        & -                                                                   & \multicolumn{1}{c|}{19.51}                                                       & -                                                                   & 6.17M                                                                                           \\
\rowcolor[HTML]{EFEFEF} 
\textbf{GoogleNet}                                                                                  & \textbf{Type-I}                                                                                        & \multicolumn{1}{c|}{\cellcolor[HTML]{EFEFEF}4.9}                                 & 0.32\%                                                              & \multicolumn{1}{c|}{\cellcolor[HTML]{EFEFEF}20.79}                               & 1.59\%                                                              & 31.49\%                                                                                         \\
\textbf{}                                                                                           & \textbf{Type-IIA}                                                                                      & \multicolumn{1}{c|}{4.83}                                                        & 0.25\%                                                              & \multicolumn{1}{c|}{21.88}                                                       & \cellcolor[HTML]{FFFFFF}2.94\%                                      & 41.97\%                                                                                         \\ \hline
\rowcolor[HTML]{EFEFEF} 
\textbf{MobileNetV1}                                                                                & \textbf{Base}                                                                                          & \multicolumn{1}{c|}{\cellcolor[HTML]{EFEFEF}10.12}                               & -                                                                   & \multicolumn{1}{c|}{\cellcolor[HTML]{EFEFEF}31.68}                               & -                                                                   & 3.30M                                                                                           \\ \hline
\textbf{MobileNetV2}                                                                                & \textbf{Base}                                                                                          & \multicolumn{1}{c|}{8.51}                                                        & -                                                                   & \multicolumn{1}{c|}{28.91}                                                       & -                                                                   & 2.41M                                                                                           \\ \hline
\end{tabular}
}
\caption{\small Comparison of \%Compression w.r.t Base Models and Test Errors on CIFAR-10 and CIFAR-100 dataset; Same networks are used for both CIFAR-10 and CIFAR-100 with only the last FC layer different; \%Compression is wrt CIFAR-100}
\label{table:SotaCNNCompression}
\end{table}

\subsection{Experiment 1: Using SCF in Heavily Over-parameterised ResNet Models}
\label{sec:exp1}
In this experiment, we choose ResNet \cite{ResNet} models constructed to study behaviours of extremely deep networks on CIFAR-10. Also, ResNet \cite{ResNet} publishes their models tailored explicitly for CIFAR-10 with 20, 32, 44 and 56 number of layers along with results. This allows us to validate our modifications to their networks rigorously. We replace standard convolutional filters from all layers with bespoke SCF configurations. We also validate using CIFAR-100, a bigger dataset with more classes, to ensure an unbiased inference of results. For CIFAR-100 dataset we use the same ResNet networks  defined for CIFAR-10 and only replace the last FC layer to produce 100 class outputs.

Figure \ref{fig:Train_Val_AccuracyResNet56} displays the training and validation error for base ResNet-56 and ResNet-56 with SCF configurations on CIFAR-100 dataset. We can make two significant observations from the figure. Firstly, the validation errors of all three networks are very close, indicating that SCF networks have generalised as good as the over-parameterised base network. Secondly, the gap between training and validation errors for the base network is much larger than SCF networks. This clearly indicates that SCF networks show less overfitting than the base network.

Figure \ref{fig:validationErrorVsParamResNet:first} and \ref{fig:validationErrorVsParamResNet:sec} display the validation error vs the number of parameters for the previously mentioned four base ResNet models on CIFAR-10 and CIFAR-100 respectively. On CIFAR-10 dataset, base ResNet-44 and Type-I ResNet-56 with 0.65M and 0.42M parameters, have a comparable validation error of 7.17\% and 7.24\% respectively. Similarly for base ResNet-32 and Type-I ResNet-44 with 0.46M and 0.33M parameters, error is 7.51\% and 7.48\% respectively. Moreover, when compared to base ResNet-20 with 0.27M parameters, Type-I ResNet-32 with 0.23M parameters and Type-IIA ResNet-44 with 0.22M parameters have lower validation error. Increasing the depth of the ResNet increases its expressiveness and thereby helps the network achieve higher accuracy. SCF allows room for ResNet to grow deeper and perform better. The trends are similar for CIFAR-100 dataset as can be seen in Figure \ref{fig:validationErrorVsParamResNet:sec}. Thus we can strongly assert that adopting SCF allows us to explore networks with more depth while keeping them economical.

\begin{figure}[] 
\centering 
\includegraphics[width=2.8in, height=1.55in]{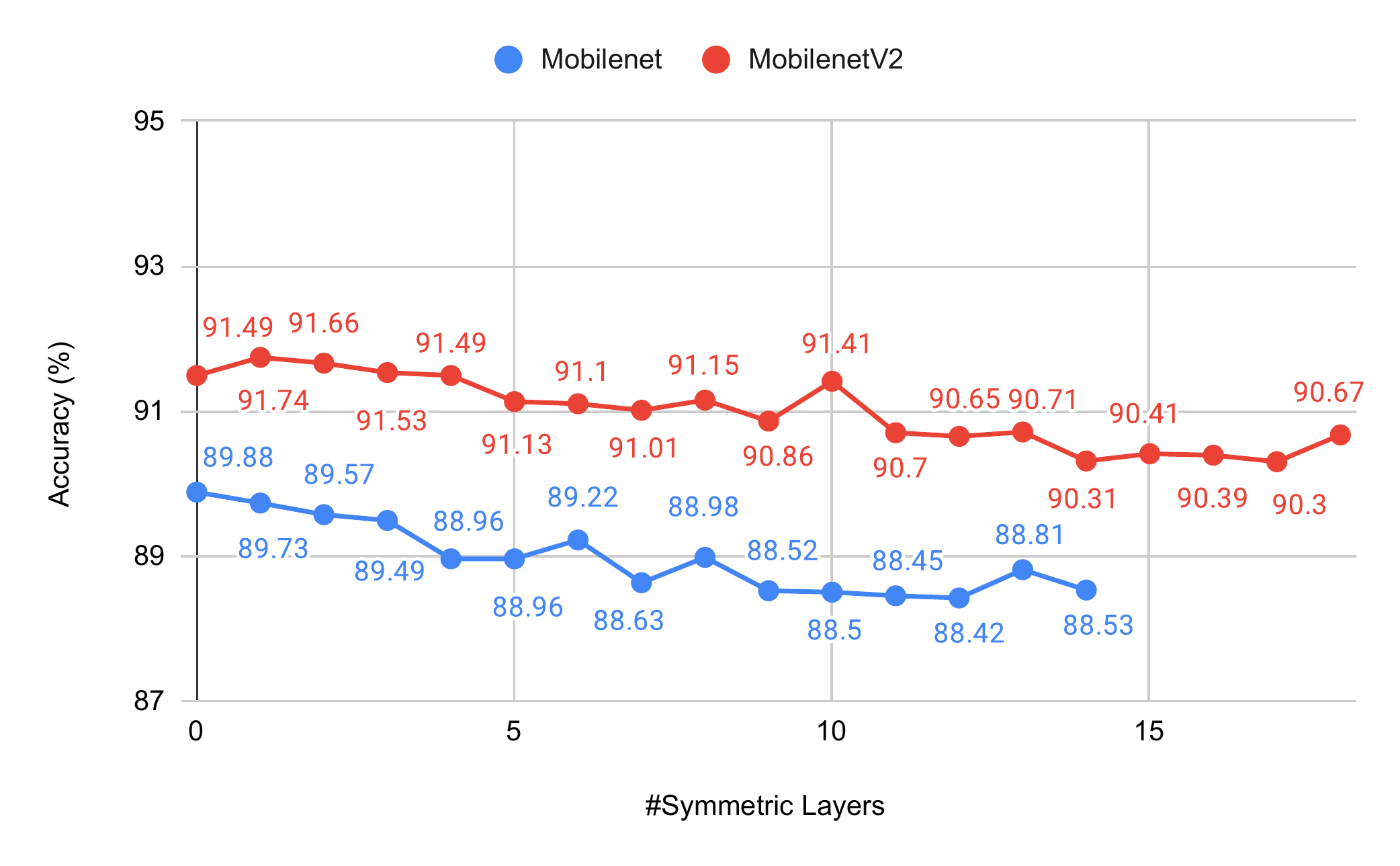} 
\caption{\small Validation Accuracy when replacing \textit{depthwise convolution} with \textit{symmetric convolution} layer by layer on MobileNets. Trained on CIFAR-10 dataset.}
\label{fig:MobileNet} 

\end{figure}

\subsection{Experiment 2: SOTA CNN Model Compression Using SCF}
\label{sec:exp2}
In this experiment, we choose two SOTA CNN models, DenseNet\cite{DenseNet} and GoogleNet\cite{GoogleNet}, having completely different basic modules, dense block and inception block respectively. Due to the presence of $1\times1$ bottleneck convolution layers, they are not only less compressible with SCF than residual blocks in ResNet but also more compact (less wide), DenseNet more so due to the dense concatenations to subsequent layers. Here again, we replace all standard convolutional filters with bespoke SCF configurations, allowing us to study the tradeoffs between accuracy and model compression. Table \ref{table:SotaCNNCompression} shows \%compression along with \%increase in test errors with respect to Base networks. Using SCF brings down their parameter comparable to \textit{edge} networks and only slightly degrades their accuracy with respect to their Base networks, yet are higher than \textit{edge} networks. 

\subsection{Experiment 3: Exploring SCF in Low Resource Edge Networks}
\label{sec:exp3}
In this experiment, we choose MobileNet V1\cite{MobileNet}, V2\cite{MobileNetV2} which are \textit{edge} networks. Although constraining the depthwise filters with symmetric filters leads to minimal compression gain (to the tune of 1-2\%), this experiment strengthens our insight into SCF in NN models. As both models are already compact, we expect that further constraining the parameters could only lead to highly degraded accuracy. To investigate, we spatially constrain the depthwise filters with Type-I symmetric filters, one layer at a time and train them on CIFAR-10 dataset. From the Figure \ref{fig:MobileNet}, we see that the validation accuracy reduces minimally for both MobileNetV1 and MobileNetV2 when symmetric filters replace all $3\times3$ depthwise filters. Contrary to our expectations, the validation accuracy of every point in the graph, representing models with varying amounts of symmetric filters, does not differ much from the base model. Remarkably, we can infer that the edge models considered optimal in parameters still have spatial redundancy that can be exploited using symmetric filters.

\subsection{Comparison of ResNet-56 Model with our SCF Configurations and Pruning Methods in Table below}
\label{sec:exp_prune}

\begin{table}[h]
%\scriptsize
%\tiny
\centering
\resizebox{\columnwidth}{1.2cm}{
\tiny
\begin{tabular}{l|c|c}
\hline
\textbf{Method}         & \textbf{CIFAR-10 Val\_Error} & \textbf{\%Original Param} \\ \hline
\rowcolor[HTML]{EFEFEF} 
ResNet-56\cite{ResNet}               & 6.97                         & 100                       \\ \hline
L1\cite{PruningEfficientCNN}                       & 6.94                         & 85.90                     \\ \hline
\rowcolor[HTML]{EFEFEF} 
HRank\cite{HRank}                    & 6.48                         & 83.20                     \\ \hline
NISP\cite{NISP}                     & 6.99                         & 57.60                     \\ \hline
\rowcolor[HTML]{EFEFEF} 
\textbf{Type-I (Ours)}   & 7.24                         & 50.28                     \\ \hline
\textbf{Type-IIA (Ours)} & 8.38                         & 33.70                     \\ \hline
\rowcolor[HTML]{EFEFEF} 
HRank\cite{HRank}                    & 10.75                        & 28.40                     \\ \hline
\end{tabular}
}
%\caption{\small Comparison of ResNet-56 model with our SCF configurations and pruning techniques from Section \ref{sec:relatedWork}}
\label{table:PruningTech}
\end{table}

\section{Conclusion}
\label{sec:conclusion}
We proposed a novel technique to constrain parameters in CNN based on symmetric filters. We investigated the impact on accuracy for CIFAR-10 and CIFAR-100 datasets when varying the combinations and levels of symmetricity in the diverse basic blocks of NN models. We believe the trends will be similar for ImageNet dataset.  We demonstrated that our models offer effective generalisation and a structured elimination of redundancy in parameters. We concluded by comparing our method with other pruning techniques. 

%\bibliographystyle{IEEEbib}
%\bibliography{strings,refs}

\end{document}